\documentclass[runningheads]{llncs}
\usepackage{graphicx}

\newcommand\blfootnote[1]{%
  \begingroup
  \renewcommand\thefootnote{}\footnote{#1}%
  \addtocounter{footnote}{-1}%
  \endgroup
}

\begin{document}

\title{Deep Simultaneous Optimisation of Sampling and Reconstruction for Multi-contrast MRI}

\titlerunning{ISMRM2020 Poster \#3619}

\author{Xinwen Liu\inst{1} \and Jing Wang\inst{1,2} \and Fangfang Tang\inst{1} \and Shekhar S. Chandra\inst{1} \and \\Feng Liu\inst{1} \and Stuart Crozier\inst{1}}
\authorrunning{X. Liu et al.}
\institute{School of Information Technology and Electrical Engineering,\\ the University of Queensland, Australia\\
\and School of Information and Communication Technology,\\ Griffith University, Australia\\
\email{feng@itee.uq.edu.au}
}

\maketitle 

\begin{abstract}
MRI images of the same subject in different contrasts contain shared information, such as the anatomical structure.  Utilizing the redundant information amongst the contrasts to sub-sample and faithfully reconstruct multi-contrast images could greatly accelerate the imaging speed, improve image quality and shorten scanning protocols. We propose an algorithm that generates the optimised sampling pattern and reconstruction scheme of one contrast (e.g. T2-weighted image) when images with different contrast (e.g. T1-weighted image) have been acquired. The proposed algorithm achieves increased PSNR and SSIM with the resulting optimal sampling pattern compared to other acquisition patterns and single contrast methods. 
\end{abstract}

\section{Introduction}
For diagnosis purposes, MRI protocols routinely involve acquiring different contrast images, such as T1-weighted images (T1WI) and T2-weighted images (T2WI). The prolonged sequential acquisition of multiple contrast images can lead to motion artefacts and discomfort for patients being scanned.  Since different contrast images of the same subject contain similar anatomical structure, the complimentary imaging information from one contrast (e.g. T1WI) could be used as a reference to sub-sample and reconstruct images with the other contrast (e.g. T2WI), thereby accelerating the overall process. Previous studies \cite{xiang2018deep} have shown the benefits of incorporating contrast information in sub-sampled reconstructions. However, most of these studies focus only on the reconstruction process, and limited investigations have been undertaken on how the complementary information of the other contrasts could be best used in the acquisition process. It is also seen from the other studies \cite{bahadir2019adaptive,bahadir2019learning} that optimisation of the sampling pattern leads to superior image quality. In this work, we take the reconstruction of sub-sampled T2WI to show that the acquisition pattern could be optimised with information provided by T1WI in order to achieve better reconstruction results. 
\blfootnote{Presented at ISMRM 28th Annual Meeting \& Exhibition (Poster \#3619)}
\section{Methods}

To learn the optimised sub-sampling pattern and reconstruction for multi-contrast images, we propose an integrated acquisition and reconstruction network, namely AR-net.  The AR-net architecture contains two parts, the acquisition network, which is used to optimise the sampling pattern, and the reconstruction network, which is based on U-net \cite{ronneberger2015u} to reconstruct the fully-sampled images from the aliased images. The detailed configuration of the network is listed in Table \ref{fig1}.

\begin{table}
  \caption{AR-net configurations.}
  \label{tbl:excel-table}
  \includegraphics[width=\textwidth]{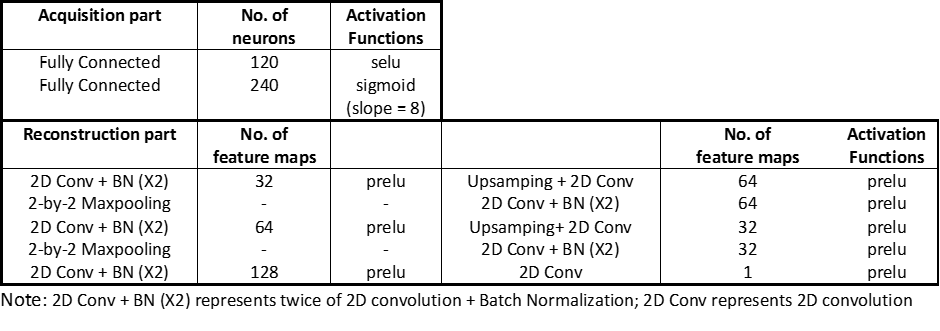}\label{fig1}
\end{table}

\begin{figure}[h]
\includegraphics[width=\textwidth]{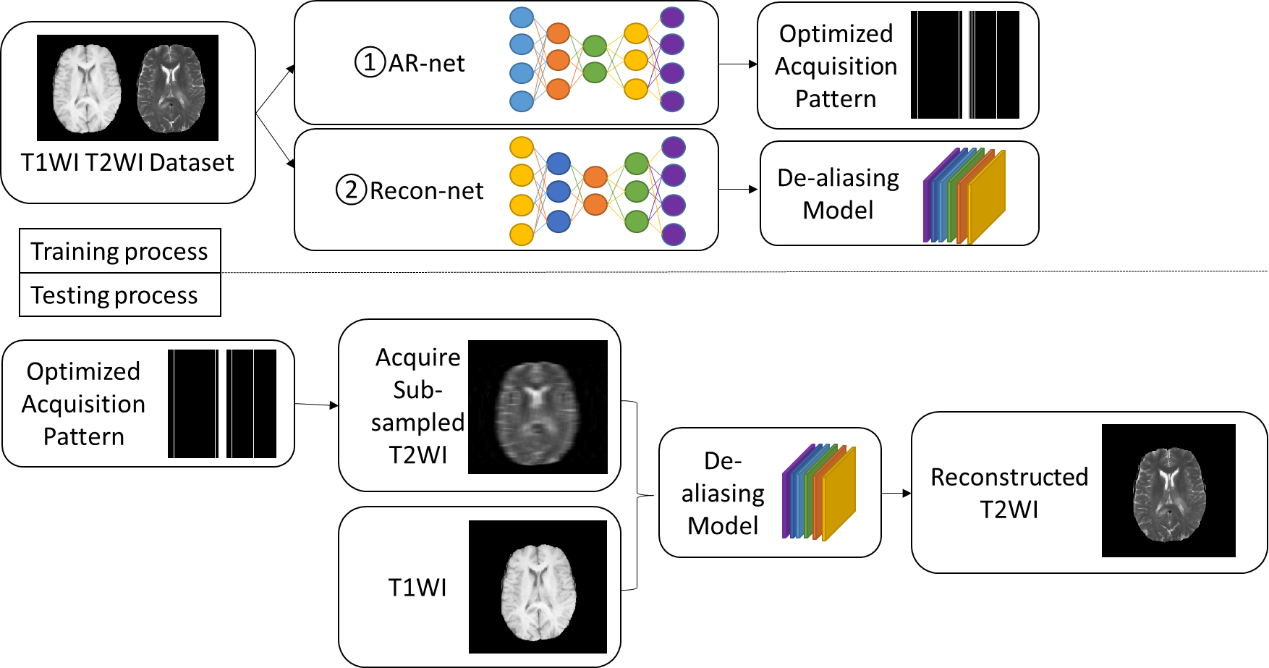}
\caption{Pipeline of the algorithm.} \label{fig2}
\end{figure}

The overall pipeline of the training and testing process is shown in Fig.\ref{fig2}. In the training process, pairs of T1WI and T2WI are firstly fed into the AR-net to train for the optimised sampling pattern. The objective is to select the most important sampling locations in k-space that lead to the minimal mean-absolute-error (MAE) between the reconstructed images and the ground truth images during the learning process. We can obtain the probability of k-space locations from the acquisition network, where higher probability represents the points to be sampled. Inspired by \cite{bahadir2019adaptive}, we use the sigmoid with large slope as the activation function and $\ell_1$-norm sparsity constraint to approximate the discrete sampling process. The slope and the coefficient of $\ell_1$-norm in the loss function are used to tune different sampling rates. The optimised mask only contains ones for sampled locations and zeros otherwise after threshold approximation. We show experiments with the 1D sub-sampling pattern optimization and it is straightforward to extend to higher dimensional patterns.

The second step in the training process is to generate the de-aliasing model that maps the zero-padded sub-sampled image to the reconstructed image using a multi-contrast reconstruction network, namely Recon-net. For simplicity, the configuration of the Recon-net is the same as the reconstruction network of AR  in Table \ref{fig1}. We simulate the sub-sampled T2WI by applying the mask generated from AR-net to the ground truth images. To further use the complementary information from the other contrast in step two, T1WI is concatenated to the aliased T2WI in the input layer of the Recon-net. We use MAE as the loss function and Adam optimiser with learning rate 0.0005 to train the model  in the training process. In the testing process, we reconstruct the T2WI which are subsampled by the learned optimised masks on the test set using the trained recon-net. 

\section{Results \& Discussion} 
We used the brain tumour segmentation challenge 2019 (BraTS2019) dataset \cite{menze2014multimodal,bakas2017advancing,bakas2018identifying,bakas2017segmentation,bakas2017segmentation1} to train and test the proposed algorithm. The data had been pre-processed by the organizer, and we further normalized the intensity to [$0\textendash1$]. We extracted 2D slices of size $240\times240$ from 208 subjects, where 2,287 slices were randomly selected as training dataset, 1,320 as validation dataset and 2,165 slices as testing dataset. The training and validation dataset are used in the training process, and the testing dataset is used only in the testing process to report the average results. 

\begin{figure}
\includegraphics[width=\textwidth]{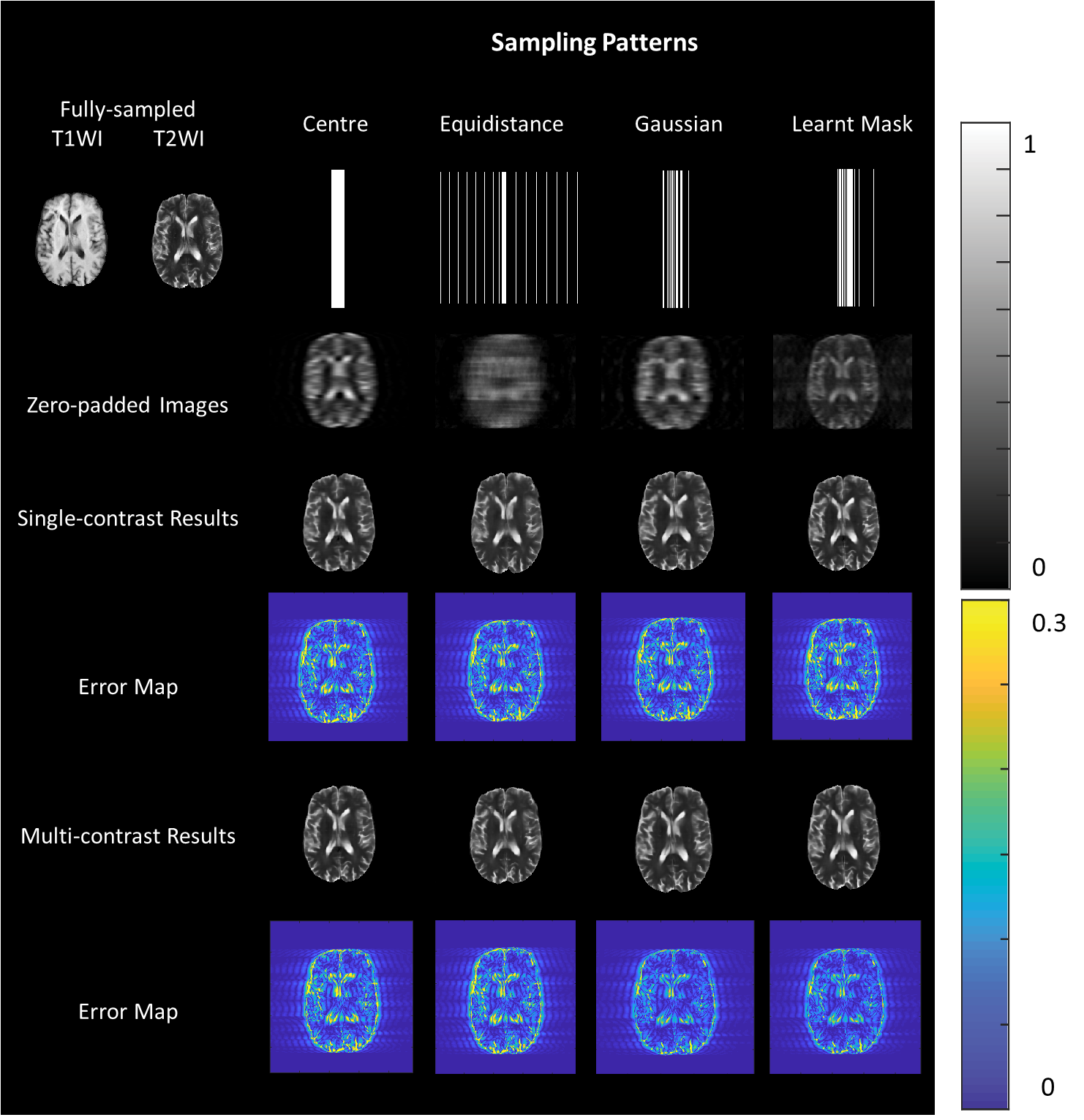}
\caption{Illustration of sampling patterns, zero-padded images, reconstructed results and the error map for acceleration rates up to 10 (22 over 240
sampled).} \label{fig3}
\end{figure}

To demonstrate the superiority of the learnt sampling pattern over conventional sampling patterns, we trained de-aliasing models for different sampling patterns, including low-resolution mask (centred sampling pattern), equal distance mask with $\frac{2}{3}$ in the centre and Gaussian mask with the same sub-sampling rate and same hyper-parameter settings. We sampled 22 over 240 lines, which is up to the sub-sampling rates of 10. The illustration of the masks and corresponding zero-padded images are shown in Fig.\ref{fig3}. To show the effectiveness of the T1WI, we also trained single-contrast de-aliasing models for T2WI and compared the results with the multi-contrast reconstruction models for different sampling patterns.

Peak signal-to-noise ratio (PSNR) and structure similarity index (SSIM) are used to evaluate the reconstruction results in the testing process. Fig.\ref{fig3} visualises the reconstruction results for different sampling patterns, and Table \ref{fig4} shows the quantitative comparison. We can observe that the proposed method yields a reduced error map with increased SSIM and PSNR despite large reduction factors. 

\begin{table}
  \caption{Comparison results between different sampling masks for single-contrast and multi-contrast reconstruction.}
  \label{tbl:excel-table}
  \includegraphics[width=\textwidth]{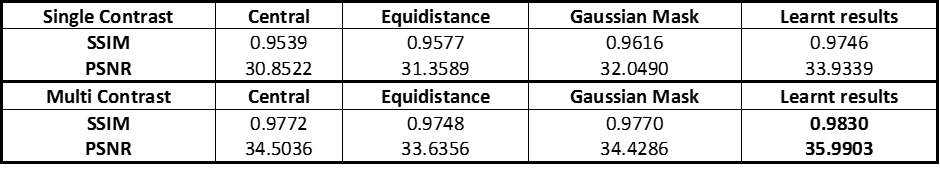}\label{fig4}
\end{table}

\section{Conclusion}
We proposed a deep learning-based AR-net that generates the optimal sampling pattern for the contrast-of-interest image and the de-aliasing model. The proposed method achieves average PSNR over 35dB and SSIM over 0.98 with acceleration rates up to 10.

\bibliographystyle{splncs}
\bibliography{main}

\end{document}